\PassOptionsToPackage{table,xcdraw}{xcolor}
\documentclass[]{fairmeta}
\usepackage[table,xcdraw]{xcolor}
\usepackage{wrapfig}
\usepackage{mathpazo}
\usepackage{tgpagella}
\usepackage{bbding}
\usepackage{graphicx}
\usepackage{makecell}
\newcommand{\ours}{MarDini}  %
\usepackage{enumitem}
\definecolor{myred}{RGB}{182, 96, 79}
\definecolor{myblue}{RGB}{98, 126, 149}
\usepackage{xfrac}
\newcolumntype{L}[1]{>{\raggedright\let\newline\\\arraybackslash\hspace{0pt}}m{#1}}
\newcolumntype{C}[1]{>{\centering\let\newline\\\arraybackslash\hspace{0pt}}m{#1}}
\newcolumntype{R}[1]{>{\raggedleft\let\newline\\\arraybackslash\hspace{0pt}}m{#1}}
\newcolumntype{Y}{>{\centering\arraybackslash}X}

\title{\ours{}: Masked Autoregressive Diffusion for Video Generation at Scale}
\author[1,2,*]{Haozhe Liu}
\author[1, \dagger]{Shikun Liu}
\author[1, \dagger]{Zijian Zhou}
\author[1]{Mengmeng Xu}
\author[1]{Yanping Xie}
\author[1]{Xiao Han}
\author[1]{Juan C. Pérez}
\author[1]{Ding Liu}
\author[1]{Kumara Kahatapitiya}
\author[1]{Menglin Jia}
\author[1]{Jui-Chieh Wu}
\author[1]{Sen He}
\author[1]{Tao Xiang}
\author[2]{Jürgen Schmidhuber}
\author[1]{Juan-Manuel Pérez-Rúa}
\affiliation[1]{Meta AI}
\affiliation[2]{KAUST}

\contribution[*]{Work done at Meta}
\contribution[\dagger]{Equal Contribution}

\abstract{
We introduce \ours{}, a new family of video diffusion models that integrate the advantages of masked auto-regression (MAR) into a unified diffusion model (DM) framework. Here, MAR handles temporal planning, while DM focuses on spatial generation in an asymmetric network design: i) a MAR-based planning model containing most of the parameters generates planning signals for each masked frame using low-resolution input; ii) a lightweight generation model uses these signals to produce high-resolution frames via diffusion de-noising. \ours{}’s MAR enables video generation conditioned on any number of masked frames at any frame positions: a single model can handle video interpolation (e.g., masking middle frames), image-to-video generation (e.g., masking from the second frame onward), and video expansion (e.g., masking half the frames). The efficient design allocates most of the computational resources to the low-resolution planning model, making computationally expensive but important spatio-temporal attention feasible at scale. \ours{} sets a new state-of-the-art for video interpolation; meanwhile, within few inference steps, it efficiently generates videos on par with those of much more expensive advanced image-to-video models.
    
}

\date{\today}
\correspondence{Haozhe Liu at \email{haozhe.liu@kaust.edu.sa}, Juan-Manuel Pérez-Rúa at \email{jmpr@meta.com}}

\metadata[Blogpost]{\url{https://mardini-vidgen.github.io}}

\begin{document}

\maketitle

\section{Introduction}
\label{section:intro}

Auto-regressive (AR) transformers \citep{vaswani2017transformer,peng2023rwkv,schmidhuber1992learningcontrol,schlag2021linear} have recently demonstrated remarkable success in natural language processing \citep{dubey2024llama,team2023gemini,achiam2023gpt}, sparking efforts to achieve similar breakthroughs in computer vision \citep{rombach2022high,dai2023emu,saharia2022photorealistic}. However, unlike the {\it discrete, sequential, and easily tokenized} nature of language, visual data consist of {\it continuous} pixel signals distributed across a {\it high-dimensional} space, making them more difficult to model through 1D auto-regression.

To overcome this challenge, recent studies have explored vector quantization techniques \citep{van2017neural,razavi2019vqvae2} to convert continuous pixel data into discrete representations suitable for AR modelling. Unfortunately, these approaches \citep{yu2022scaling,ramesh2021zero} rely on {\it causal attention}, which is not well aligned for high-dimensional visual data, often leading to diminished performance \citep{li2024autoregressive}, particularly on large-scale datasets \citep{xie2024showo, zhou2024transfusion}. To mitigate this limitation, masked auto-regression (MAR) has been introduced \citep{chang2022maskgit, li2023mage}. MAR replaces the causal attention with {\it bi-directional attention} \citep{he2021mae,devlin2019bert}, effectively simulating auto-regressive behaviour while being more capable of handling visual data. Leveraging this approach, MAR exhibits flexibility in handling diverse generation tasks through different masking strategies, such as image generation \citep{chang2022maskgit, li2023mage}, out-painting \citep{chang2022maskgit}, video expansion \citep{yu2023magvit} and class-conditioned video generation \citep{yu2023language,voleti2022mcvd} while maintaining manageable computational overhead.  Although MAR shows potential in scaling image and video generation tasks \citep{chang2023muse,yu2023magvit,yu2023language}, its key bottleneck lies in its training instability which is tied to the reliance on discrete representations \citep{ramesh2021zero, razavi2019vqvae2}.

Meanwhile, Diffusion models (DMs) \citep{ho2020denoising,neal2001annealed,jarzynski1997equilibrium} have emerged as a successful alternative for scaling vision generative models, offering stable training by modelling visual signals directly in a continuous space. However, DMs tend to incur high inference costs due to the requirement of the multi-step diffusion process. Here, video generation poses an even greater challenge --- Video is a {\it strict super-set} of the image domain, requiring additional modelling for temporal consistency and complex motion dynamics. 

To this end, we propose a new paradigm for video generation that combines the flexibility of MAR in a {\it continuous} space with the {\it robust generative capabilities} of DM. Specifically, we present a scalable training recipe and an efficient neural architecture design for video generation. Our model decomposes video generation into two sub-tasks --- temporal and spatial modelling ---  handled by {\it distinct} networks with {\it an asymmetric design} based on the following two principles:

\begin{enumerate}[leftmargin=1.2em]
    \item {\it MAR handles long-range temporal modelling, while DM focuses on detailed spatial modelling.}
    \item {\it MAR operates with more parameters at a lower resolution, while DM operates with fewer parameters at a higher resolution.}
\end{enumerate}

Following these principles, we use the same training batch for both MAR and DM but employ two distinct processes operating at different resolutions. 
MAR receives randomly masked low-resolution input frames and predicts the corresponding planning signals.
Conditioned on these planning signals via cross-attention and the unmasked frames, DM learns to incrementally recover the masked high-resolution frames from noise. Finally, we introduce a progressive training strategy that gradually curates mask ratios and with its data pipelines, allowing our model to be {\it trained from scratch on unlabeled video data}. This eliminates the common reliance on text-to-image and text-to-video pre-training, as seen in other video diffusion models \citep{girdhar2023emu,blattmann2023stable}.

Our model integrates MAR-based planning signals with a DiT-based \citep{peebles2023dit, chen2024gentron} lightweight, tiny diffusion model, hence the name {\bf \ours{}}. Our empirical study on \ours{} highlights the following key characteristics:

\begin{itemize}[leftmargin=1.2em]
    \item {\bf Flexibility.} With MAR conditioning, \ours{} naturally supports a range of video generation tasks through flexible masking strategies. For example, when given the first frame and masking the rest, it performs image-to-video generation; when given a video and masking subsequent frames, it performs video expansion; and, when given the first and last frames and masking the middle frames, it performs video interpolation. By hierarchically and auto-regressively masking middle frames across multiple inferences, \ours{} generates slow-motion videos.
    \item {\bf Scalability.} \ours{} can be trained from scratch at scale, without relying on generative image-based pre-training. In contrast to most video generation models, that treat video as a secondary task following image generation, \ours{} leverages mask ratio tuning to progressively adjust the difficulty of the training task. This approach enables the model to scale from video interpolation to full video generation, directly bypassing the need for image-based pre-training.
    \item {\bf Efficiency.} \ours{}’s asymmetric design allocates more computational resources to lower resolutions, making it memory-efficient and fast during inference. With lower overall memory usage, \ours{} allows the deployment of computationally intensive spatio-temporal attention mechanisms at scale, improving its ability to model complex motion dynamics.
\end{itemize}

\section{\ours{}: An Efficient and Asymmetric Video Diffusion Model}

\subsection{Design Overview}
\begin{figure}[t!]
    \centering
    \includegraphics[width=\linewidth]{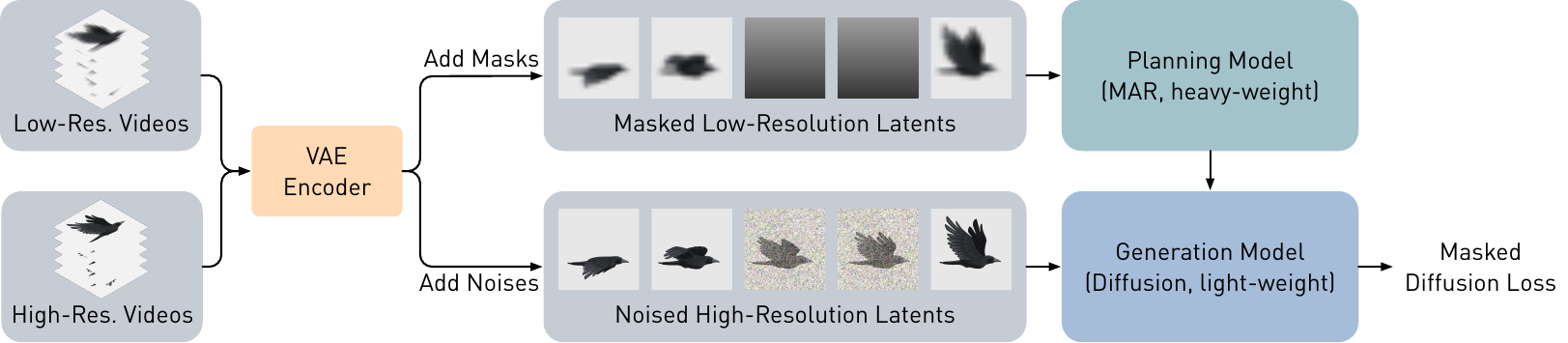}
    \caption{{\bf \ours{} Training Pipeline Overview.} A latent representation is computed for unmasked frames that serve as a conditional signal to a generative process. On the first hand, we have a planning model that autoregressively encodes global conditioning signals from a low-resolution version of the unmasked latent inputs. On the other hand, the planning signals are fed to the diffusion-based generation model through cross-attention layers. A high-resolution version of the input conditions is also ingested by the diffusion model, enabling generation with a coherent temporal structure and a direct mechanism to attend to fine-grained details of the unmasked frames. \ours{} is trained end-to-end via masked frame-level diffusion loss.}
    \label{fig:pipeline}
\end{figure}

\ours{} is a video generation model designed to efficiently generate high-resolution videos using an asymmetric network architecture. As shown in Figure~\ref{fig:pipeline}, \ours{} consists of two networks: a heavy-weight MAR \textit{planning} model and a light-weight \textit{generation} DM. During training, the planning network processes randomly masked low-resolution frames and predicts corresponding planning signals. These planning signals compress the semantic and long-range temporal information, guiding the DM's high-resolution generation process. The DM receives noisy frames at the masked positions and reconstructs them by progressively removing noise. 

In this section, we outline and address the key design challenges involved in training \ours{}. First, we describe the data representations and their corresponding notations within the \ours{} framework (Section~\ref{sec:data_representation}). Next, we describe the design details of the MAR planning network and the DM, along with the integration of additional guidance such as diffusion steps and planning signals (Section~\ref{sec:architecture_design}). Finally, we outline the multi-stage training recipe for \ours{}, which we found to be essential for ensuring stable training (Section~\ref{sec:training_recipes}). Collectively, these innovations enable \ours{} to become one of the first video generation models capable of being trained from scratch using only unlabelled video data.

\subsection{Data Representation and Notations}
\label{sec:data_representation}

\paragraph{VAE Compressor.}
Consistent with prior works \citep{dai2023emu, girdhar2023emu}, we adopt a pre-trained Variational Auto-Encoder (VAE) \citep{Kingma2013AutoEncodingVB}, denoted by $\mathcal{D}_\text{enc}$, to compress videos into a low-dimensional continuous latent space, which improves both training and inference efficiency. Our VAE employs a %
16-channel latent dimension with an 8$\times$ spatial compression rate to preserve spatial details, following \cite{dai2023emu}. The VAE outputs are then patchified into a shape of $N\times C$, where $N$ represents the token count and $C=16$ represents its latent dimension.

\paragraph{MAR Planning Model.} 
Given a low-resolution input video $\mathbf{X}_\text{low} = \{x^\text{low}_i\}_{i=1:K}$ with $K$ frames, we apply the VAE encoder to compress the frames into their corresponding latent representations: $\mathbf{Z}_\text{low} = \{ z^\text{low}_i\}_{i=1:K} = \mathcal{D}_\text{enc}(\mathbf{X}_\text{low})$. To train the MAR planning model $\mathcal{P}$, we randomly select $K'<K$ video latents $\{z^{low}_j\}_{j=1:K'} \in \mathbf{Z}_\text{low}$ and replace them with a learnable mask token \texttt{[MASK]}, resulting in the final masked low-resolution latent inputs $\mathbf{Z}_\text{low}^\text{mask}$. The planning model then processes $\mathbf{Z}_\text{low}^\text{mask}$ and predicts $\mathbf{Z}_\text{cond} = \mathcal{P}(\mathbf{Z}_\text{low}^\text{mask}) =  \{z^\text{cond}_i\}_{i=1:K}$, where $z^\text{cond}_i$ is the planning signal for the $i$-th frame, shaped as $N_\text{low}\times C_\text{low}$, with $N_\text{low}$ representing the number of patches per frame.

\paragraph{DM Generation Model.} 
Conversely, we obtain high-resolution video latents $\mathbf{Z}_\text{high} = \{ z^\text{high}_i\}_{i=1:K} = \mathcal{D}_\text{enc}(\mathbf{X}_\text{high})$ with dimensions $N_\text{high} \times C_\text{high}$, generated by the VAE encoder using the same video inputs at high resolution: $\mathbf{X}_\text{high} = \{x^\text{high}_i\}_{i=1:K}$. Notably, we have \( N_\text{high} \gg N_\text{low} \). At diffusion step $t$, we sample noise and add it to $K'$ frames that were masked in the planning model (denoted by {\tt [NOISE]}), leaving the remaining $K-K'$ reference frames unchanged (denoted by {\tt [REF]}). This produces the final noisy high-resolution video latent inputs $ \mathbf{Z}_\text{high}^{\text{noise},t}$. Then, the generation model $\mathcal{G}$ processes these latent inputs $\mathbf{Z}_\text{high}^{\text{noise},t}$ and performs a standard denoising step, where we denote the DM output at time step $t$ as $\mathcal{G}(\mathbf{Z}_\text{high}^{\text{noise},t}, \mathbf{Z}_\text{cond}, t)$.

\subsection{Architecture Design}
\label{sec:architecture_design}
In this section, we provide a comprehensive explanation of the \ours{} architecture, including its detailed design, model configurations, and variations.

\subsubsection{\ours{} Block Design}

\begin{figure}[t!]
    \centering
    \includegraphics[width=0.85\linewidth]{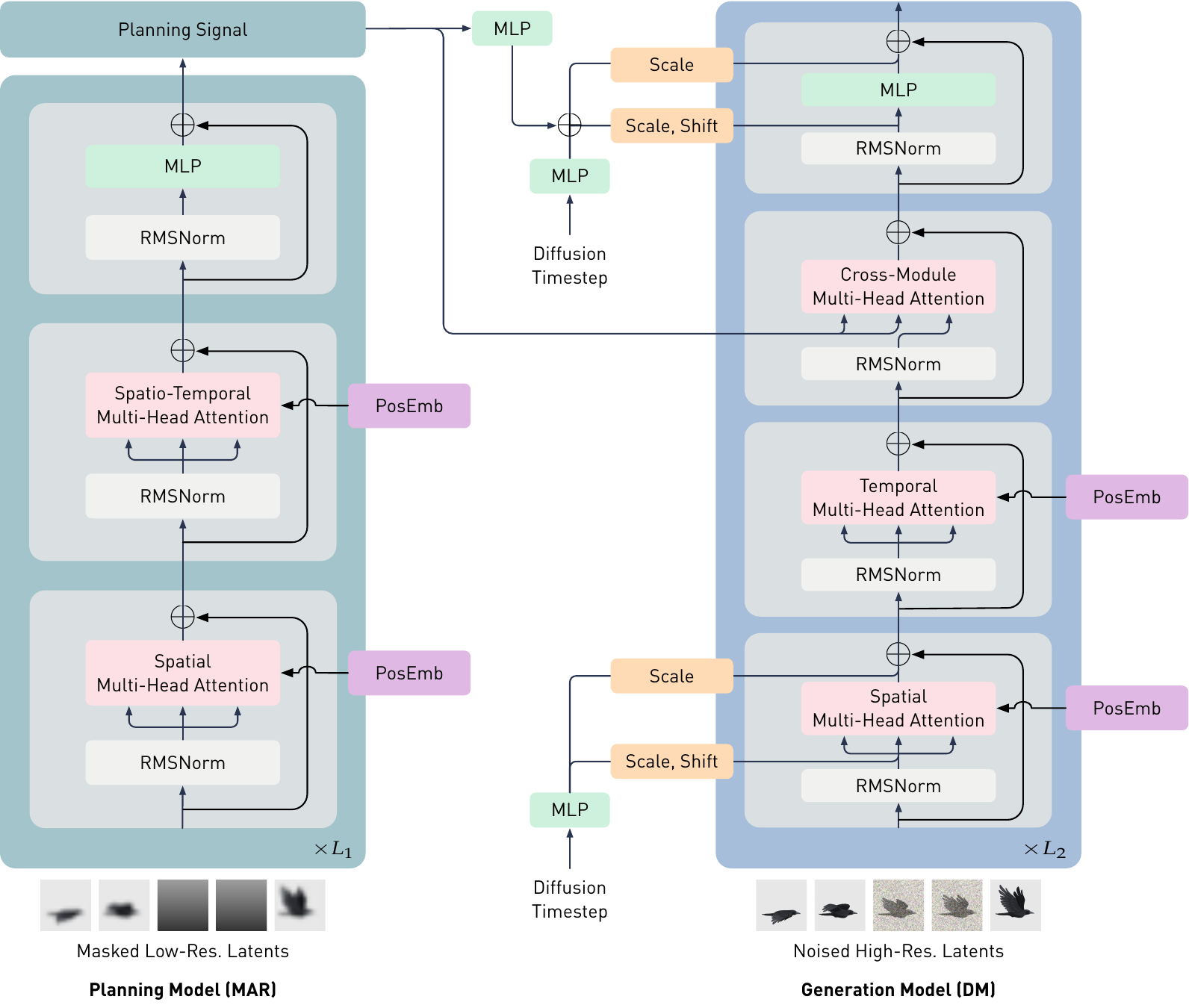}
    \footnotesize
    \caption{{\bf \ours{} Design Details.} \ours{} employs a transformer architecture for both the planning and generation models, incorporating a DiT-style block for the generation model and a Llama-style block for the planning model. We set \( L_1 \gg L_2 \), where $L_{1}$ and $L_2$ refer to the number of layers in the planning and generation model respectively. }
    \label{fig:block}
\end{figure}

Figure~\ref{fig:block} illustrates the design of the \ours{}'s MAR and DM models, both of which are based on the transformer architecture \citep{vaswani2017transformer}. 

In the MAR planning model, we adhere to the design conventions established in Llama models \citep{dubey2024llama,touvron2023llama}, which apply RMS-Norm \citep{zhang2019root} to normalize the inputs of each attention block. Additionally, layer normalization \citep{Ba2016LayerN} is applied to normalize the projected features in multi-head attention, enhancing training stability. Due to the use of low-resolution inputs, we manage to directly employ spatio-temporal attention, allowing tokens to attend across frames. This design is feasible only with asymmetric resolution inputs, as it prevents excessive memory consumption.

Concretely, within each attention block in MAR, we utilize rotary positional encoding (RoPE) \citep{su2024roformer} to encode both the spatial and temporal positions of the video tokens. To accomplish this, we apply a 2D RoPE to encode the 3-dimensional video data. Specifically, we flatten the image patches into a 1-dimensional token sequence and insert a learnable \texttt{[NEXT]} token to differentiate image patches across different rows, following \cite{gao2024lumina}. This design effectively handles video data with varying aspect ratios and resolutions.

We design the DM model in alignment with MAR, but with three key differences. First, we adopt a DiT-style approach \citep{peebles2023dit}, using AdaIN \citep{huang2017arbitrary} to integrate the diffusion steps as a conditional signal within the spatial attention layers, and additionally added with the MAR's planning signal within the MLP layers. Second, we introduce a cross-attention layer to process the planning features predicted by the MAR model. Lastly, we replace spatio-temporal attention with temporal attention \citep{blattmann2023align} to reduce the computational cost associated with high-resolution inputs in DM.

\subsubsection{Identity Attention}
\begin{wrapfigure}{r}{0.3\linewidth}
    \vspace{-1.8em}
    \centering
    \includegraphics[width=\linewidth]{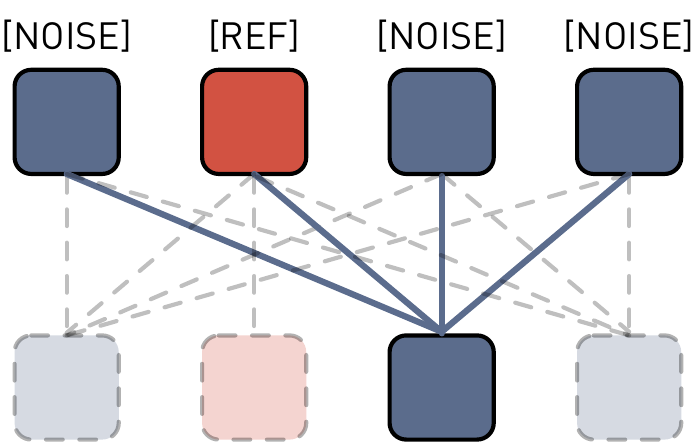}\vspace{0.2em}
    \includegraphics[width=\linewidth]{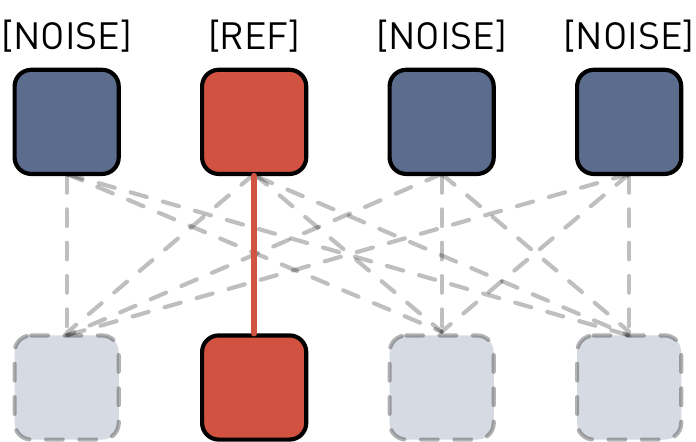}
    \caption{{\bf Identity Attention Design Details in DM.} In this setup, \texttt{[REF]} tokens only attend to themselves, while \texttt{[NOISE]} tokens attend to all other tokens across different frames. }
    \vspace{-1em}
    \label{fig:identity_attention}
\end{wrapfigure}

In our initial experiments, we observed significant training instability in \ours{}'s DM. We speculate that this is due to two main factors: i) the inherent distributional disparity between noisy (\texttt{[NOISE]}) tokens and clean reference (\texttt{[REF]}) tokens, which is further amplified by the stochastic nature of sampling diffusion steps; and ii) the random positions and varying lengths of these \texttt{[NOISE]} tokens. These factors likely compound, potentially disrupting the DM's training signals and hindering the model's ability to converge efficiently.

To address this challenge, we introduce Identity Attention, which enables the model to easily distinguish between \texttt{[REF]} and \texttt{[NOISE]} tokens by employing a separate attention strategy. As illustrated in Figure~\ref{fig:identity_attention}, \texttt{[REF]} tokens simply serve as an {\it identity projection}, preserving the input reference frames without attending to other tokens. In contrast, \texttt{[NOISE]} tokens possess a global view, attending to tokens across {\it all frames}. The \texttt{[REF]} tokens serve as guidance for generation, so we design them to be isolated from other tokens, while \texttt{[NOISE]} tokens provide global attention to all conditional signals for generation. We incorporate Identity Attention in both the spatio-temporal layers of MAR and the temporal layers of DM, which has been found to significantly enhance training stability in both models.

\subsubsection{Model Configuration}
\label{sec:model_configuration}

As outlined in Table~\ref{tab:config}, this study develops four models with distinct configurations. We train two planning models with 3.1B and 1.3B parameters alongside two generation models, employing spatio-temporal or temporal attention mechanisms. To align with our asymmetric design between the planning and generation models, the generation model's parameter size is reduced to $3\times$ or $10\times$ smaller than that of the planning model. Due to the high computational cost of spatio-temporal attention, we limit \ours{}-L/ST and \ours{}-S/ST to a 9-frame length for fair comparison on VIDIM-Bench \citep{jain2024video}. Importantly, the model's ability to autoregressively generate samples ensures that the length of the output video is not constrained.

\begin{table}[!ht]
\centering
\footnotesize
\setlength\tabcolsep{3.5pt}
\begin{tabular}{lccccccccccc}
\toprule
\multirow{2}{*}{Configuration} & \multicolumn{5}{c}{Planning Model (MAR)}                        & \multicolumn{5}{c}{Generation Model (DM)}                       & \multirow{2}{*}{Frame} \\  \cmidrule(r){2-6} \cmidrule(l){7-11}
                               & Depth & Hidden Size & MLP Size & Attn.                 & Param. & Depth & Hidden Size & MLP Size & Attn.                 & Param. &                                     \\ \midrule
\ours{}-S/ST                          & 8     & 4096        & 4096     & S.-T. Attn. & 1.3B  & 8     & 1024        & 4096     & S.-T. Attn. & 288M  & 9                                   \\
\ours{}-L/ST                         & 16    & 4096        & 8192     & S.-T. Attn. & 3.1B   & 8     & 1024        & 4096     & S.-T. Attn. & 288M  & 9                                   \\
\ours{}-S/T                           & 8     & 4096        & 4096     & S.-T. Attn. & 1.3B  & 8     & 1024        & 4096     & T. Attn.        & 288M  & 17                                  \\
\ours{}-L/T                          & 16    & 4096        & 8192     & S.-T. Attn. & 3.1B   & 8     & 1024        & 4096     & T. Attn.        & 288M  & 17 \\
\bottomrule
\end{tabular}
\caption{{\bf Configuration Details of \ours{} Models.} We provide four models, differing primarily in the size of the planning module (3.1B vs. 1.3B parameters) and the attention mechanisms used in the generation module: spatio-temporal attention (S.-T. Attn.) vs. temporal attention (T. Attn.).}
\label{tab:config}
\end{table}

\subsection{\ours{} Training Recipes}
\label{sec:training_recipes}

In this section, we outline the training pipeline of \ours{}. Specifically, we employ a multi-stage progressive training strategy that gradually increases task difficulty. This approach offers two key benefits: i) progressive learning inherently enhances training stability and improves the performance of generative models, as demonstrated by \cite{karras2017progressive} and \cite{chen2024pixart}; and ii) it allows for the collection of checkpoints from earlier stages, which helps mitigate setbacks caused by suboptimal configurations. Below, we elaborate on our detailed progressive training strategy, including the training objectives, architecture design, and training data configurations. A comprehensive training manual for \ours{} is shown in Figure~\ref{fig:recipe}, with detailed hyper-parameters and optimization methods further outlined in the Appendix \ref{app:sec:training_detials}.

\begin{figure}[!t]
    \centering
    \includegraphics[width=\linewidth]{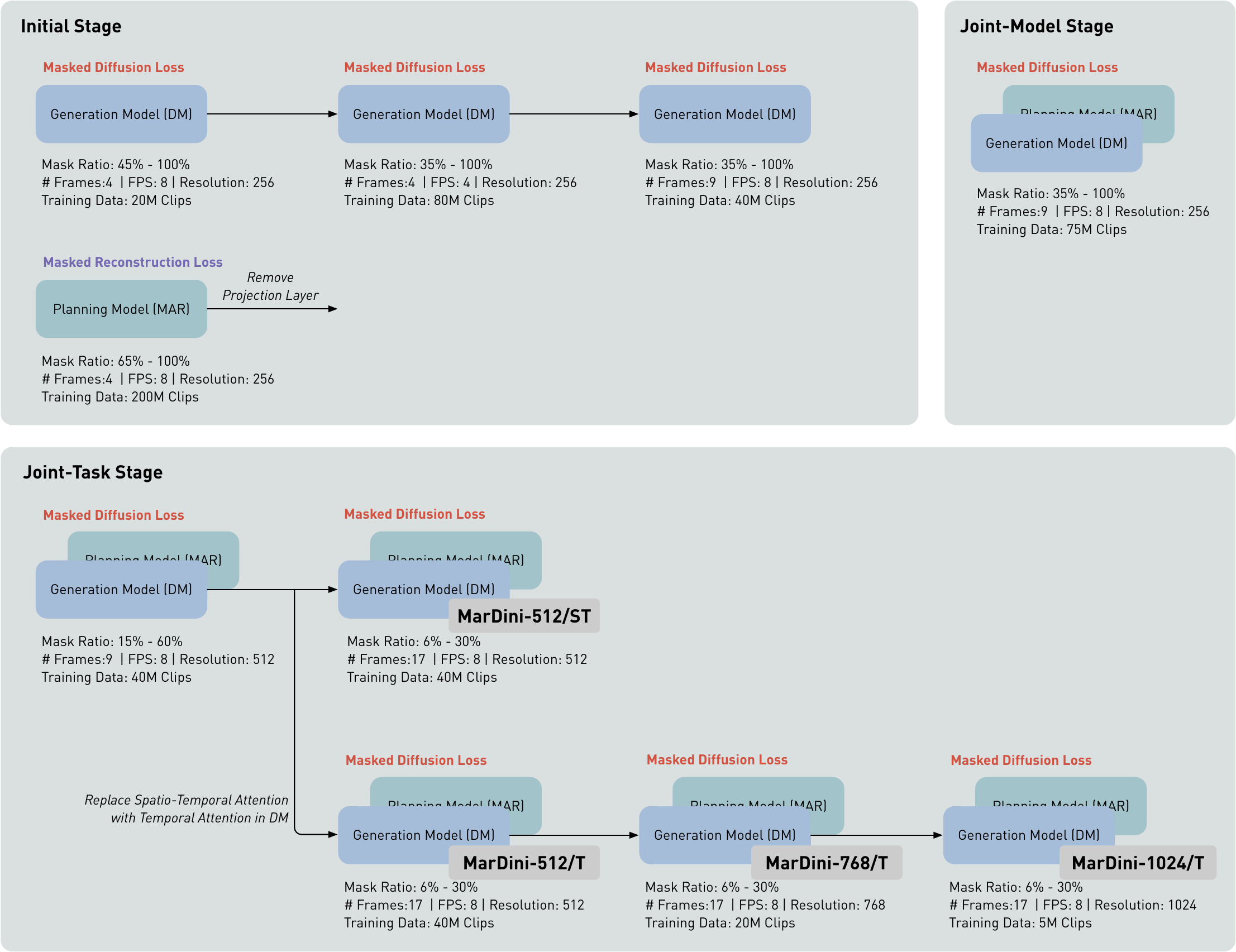}
    \caption{{\bf \ours{} Training Manual.} We list the mask ratios, frame rate (FPS), number of frames, and the size of training data for each training stage. This training manual applies to both small (\ours{}-S) and large (\ours{}-L) models. Note that the total training data refers to the amount of data observed by the model for gradient updates, rather than the vanilla size of the training dataset. Our final model checkpoints are highlighted in \textcolor{gray}{{\bf gray}}. }
    \label{fig:recipe}
\end{figure}

\subsubsection{Training Tasks: From Frame Interpolation to Video Generation}
Our training objectives are organized into three stages: i) Initial Stage: We separately train the planning and generation models, each with its own learning objective, to initialize their model weights. ii) Joint-Model Stage:  We combine the models for joint training on a simple video interpolation task, using only a masked diffusion loss. iii) Joint-Task Stage: We further train the model by gradually reducing the number of preserved reference frames, enabling it to jointly learn video interpolation and image-to-video generation tasks.

\paragraph{Initial Stage.} \cite{wang2024deepnet} pointed out that transformers with a large parameter count often experience unstable training. As such, we simplify the training dynamics by separately warming up the two models as an initial step. 

To optimize generation model $\mathcal{G}$, we employ a masked diffusion loss $\mathcal{L}_\text{DM}$: 
\begin{align}
    \mathcal{L}_\text{DM}^\theta = || \mathbf{M} \cdot \mathbf{V}^t -\mathbf{M} \cdot \mathcal{G}_\theta(\mathbf{Z}_\text{high}^{\text{noise}, t}, \mathbf{Z}_\text{uncond}, t) ||_2^2,
\end{align}
where \( \mathbf{Z}_\text{uncond} \) is a learnable token serving as {\it unconditional guidance} from the planning model. \( \theta \) represents the parameters of the generation model, and \( \mathbf{M} \) denotes the binary masks used to mask out all clean reference frames. Inspired by \cite{blattmann2023align, salimans2022progressive}, we apply velocity prediction as the diffusion loss, where the prediction target \( \mathbf{V}^t = \{v_i^t\}_{i=1:K} \) represents the velocity at time step \( t \) for the \( i \)-th frame, defined as \( v_i^t = \alpha_t \epsilon - \sigma_t z^\text{high}_i , \epsilon \sim \mathcal{N}(0,I) \). Here, \( \alpha_t \) and \( \sigma_t \) correspond to the diffusion scheduler at \( t \) step.

To optimize MAR planning model $\mathcal{P}$, we employ a masked reconstruction loss $\mathcal{L}_\text{MAR}$:
\begin{align}
   \mathcal{L}^{\phi, \zeta}_\text{MAR} = || \mathbf{M} \cdot \mathbf{Z}_\text{low} -  \mathbf{M} \cdot f_\zeta(\mathcal{P}_\phi(\mathbf{Z}_\text{low}^\text{mask}) ||_2^2.
\end{align}
where $f$ denotes a projection layer that depatchifies the model predictions to match the resolution of the low-resolution input image $\mathbf{Z}_\text{low}$. $\phi, \zeta$ represent the learnable parameters of the planning model and the projection layer respectively. Note that, $f$ is only used during the initial training stage, and will be removed in the later training stages.

\paragraph{Joint-Model Stage.}
After the initial pre-training stage, we then jointly train the planning and generation models end-to-end using a unified masked diffusion learning objective $\mathcal{L}_\text{MDiff}$:
\begin{align}
    \mathcal{L}_\text{MDiff}^{\theta,\phi} = || \mathbf{M} \cdot \mathbf{V}^t -\mathbf{M} \cdot \mathcal{G}_\theta(\mathbf{Z}_\text{high}^{\text{noise}, t}, \mathcal{P}_\phi(\mathbf{Z}_\text{low}^\text{mask}), t) ||_2^2,
\end{align}
where $\mathbf{Z}_\text{cond} = \mathcal{P}(\mathbf{Z}_\text{low}^\text{mask})$ is the planning signal predicted by MAR. In order to enable classifier-free guidance \citep{ho2022classifier} on the planning signal, we maintain a fixed probability of \sfrac{1}{10} to randomly replace $\mathbf{Z}^t_\text{cond}$ with $\mathbf{Z}_\text{uncond}$.

\paragraph{Joint-Task Stage.} 
In the final training stage, we reuse the learning objective from the previous stage, but gradually decrease the masking ratio to induce more challenging generation tasks. Here, mask ratio refers to the proportion of frames preserved during training. This stage requires a significantly larger computational resources with higher-resolution videos, as it determines the model's final performance. By gradually decreasing the masking ratios, we smoothly transform the model's task from video interpolation to single-image-to-video generation. This procedure ultimately enables the model to generate videos with a variable number of input frames at arbitrary temporal locations.

\subsubsection{DM Architecture: From Spatio-Temporal to Temporal Attention}
In conjunction with our progressive training objectives, we also introduce a progressive architectural design. Specifically, we first use spatio-temporal attention in the DM during the initial training stage. This choice promotes convergence, compared to temporal attention, as noted in \cite{gao2024lumina}. Since in our initial stage we train the DM in isolation and on a relatively low-resolution setup, this sophisticated attention incurs in minor computational overhead. When integrating MAR with the DM in the second stage, we replace the spatio-temporal attention with the more cost-effective temporal attention, thus increasing the efficiency of the generation model.

\subsubsection{Data: Progressive Configuration of Specifications}
Analogous to our progressive strategies for training objective and architecture we also propose a progressive data configuration. Over time, we gradually increase the video's spatial resolution, alongside progressively extending the video's duration. This approach ensures efficient use of computational resources and facilitates effective model scaling, allowing \ours{} to handle more complex and high-resolution video data as training progresses.

\section{Experiments}

We evaluate \ours{} on two benchmarks: VIDIM-Bench \citep{jain2024video}, for long-term video interpolation, and VBench \citep{huang2023vbench} for image-to-video generation. We further elaborate on the specifics of these benchmarks in Appendix \ref{app:sec:bench}. {\it We highly encourage referring to the generated videos in our web page for a comprehensive understanding of the quality of the generated videos.}

\subsection{Ablation Studies and Analysis}

\paragraph{Effectiveness of MAR and DM.}

We demonstrate the importance of having a DM on top of our MAR planning model. In fact, it is tempting to hypothesize that MAR on its own contains all the ingredients to enable high-quality video interpolation. To explore this, we introduce a projection layer to directly unpatchify the output of the MAR model without intermediate diffusion. Our experiments on VIDIM-Bench reveal that, MAR on its own, performs poorly on interpolation tasks, as shown by the first two and last two rows in Table~\ref{table:ab}, for both the 1B and 3B settings. This result suggests that directly applying MAR to continuous space is suboptimal, a result consistent with previous findings \citep{li2024autoregressive}. Similarly, directly tackling this task with a small DM without global guidance, according to the third row of  Table~\ref{table:ab}, results in sub-optimal performance. However, by combining MAR's planning capability with DM's stable performance in continuous space, we achieve optimal results, demonstrating that both components are beneficial for video generation.

\begin{table}[!ht]
\begin{minipage}{0.4\textwidth}
\centering
\caption{{\bf Effectiveness of MAR and DM design.} The reported results are FVD on VIDIM-Bench. All experiments are evaluated at a resolution of $[256 \times 256]$ using DDIM scheduler with 25 steps.}
\label{table:ab}
\footnotesize
\setlength{\tabcolsep}{1em}  
\renewcommand{\arraystretch}{1.14}
\resizebox{.99\textwidth}{!}{
\begin{NiceTabular}{cccc}
\toprule
\multirow{2}{*}[-2pt]{\makecell{Planning \\ Model}} & \multirow{2}{*}[-2pt]{\makecell{Generation\\ Model}} & \multicolumn{2}{c}{FVD$\downarrow$} \\ \cmidrule(l){3-4}
& & DAVIS & UCF101 \\ \midrule 
MAR-1B          & -                 & 427.66           & 741.80   \\ 
MAR-3B          & -                 & 373.03           & 701.03   \\ 
    \midrule 
-               & DM-0.3B           & 320.89           & 383.04   \\ 
    \midrule 
MAR-1B          & DM-0.3B           & \cellcolor{metabg}224.07           & \cellcolor{metabg} 258.08   \\ 
MAR-3B          & DM-0.3B           & \cellcolor{metabg}\textbf{102.87}  & \cellcolor{metabg}\textbf{197.69}   \\ 
\bottomrule
\end{NiceTabular}
}
\end{minipage}\hfill
\begin{minipage}{0.56\textwidth}
\centering
\caption{{\bf Efficiency of the \ours{}'s generations with and without the asymmetric design.} Both latency and GPU memory is measured as the average time to generate a video using DDIM with 25 steps using a single A100 GPU, and with {\tt bf16} mixed precision. 
}
\label{tab:efficiency}
\footnotesize
\setlength{\tabcolsep}{0.5em}  
\renewcommand{\arraystretch}{0.89}
\setlength\tabcolsep{2pt}
\resizebox{.99\textwidth}{!}{
\begin{NiceTabular}{ccccccc}
\toprule
\multirow{2}{*}[-2pt]{\makecell{Asymm. \\ Attention}} & \multirow{2}{*}[-2pt]{\makecell{Asymm. \\ Resolution}} & \multirow{2}{*}[-2pt]{\makecell{\# Inference \\ Frames}} & \multicolumn{2}{c}{[256 $\times$ 256]}                        & \multicolumn{2}{c}{[512 $\times$ 512]}       \\ \cmidrule(r){4-5} \cmidrule(l){6-7}
                            &                            &                         & Latency                 & GPU Mem.                 & Latency         & GPU Mem.        \\ \midrule
\XSolidBrush                           & \XSolidBrush                           & 9 (1 to 8)              & \multirow{2}{*}{2.76 s} & \multirow{2}{*}{25.22 G}  & 25.09 s         & 74.44 G         \\
\XSolidBrush                          &          \Checkmark                 & 9 (1 to 8)              &                         &                          & 17.91 s         & 41.03 G         \\ \midrule
\XSolidBrush                          & \XSolidBrush                           & 13 (1 to 12)            & \multirow{2}{*}{4.41 s} & \multirow{2}{*}{27.80 G} & \multicolumn{2}{c}{Out of Memory} \\
\XSolidBrush                           & \Checkmark                          & 13 (1 to 12)            &                         &                          & 34.58 s         & 62.51 G         \\ \midrule
\Checkmark                           & \XSolidBrush                         & 13 (1 to 12)            & \multirow{2}{*}{\textbf{2.63 s}} & \multirow{2}{*}{\textbf{27.75 G}} & \multicolumn{2}{c}{Out of Memory} \\
\Checkmark                           & \Checkmark                          & 13 (1 to 12)            &                         &                          & \textbf{6.05 s}          & \textbf{42.57 G}        \\
\bottomrule
\end{NiceTabular}
}
\end{minipage}
\end{table}

\begin{figure}[ht!]
    \centering
    \begin{subfigure}[t]{0.28\textwidth}
        \centering
          \includegraphics[width=\linewidth]{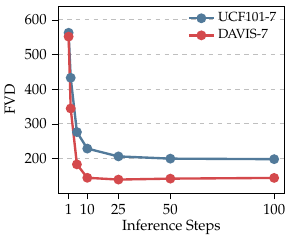}
        \caption{\footnotesize Video interpolation results with varying inference steps.}
        \label{subfig:ablation_a}
    \end{subfigure}\hfill
    \begin{subfigure}[t]{0.362\textwidth}
        \centering
        \includegraphics[width=\linewidth]{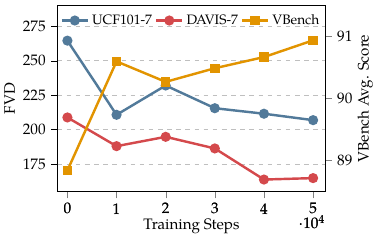}
        \caption{\footnotesize Relationship between video interpolation and image-to-video generation.}
        \label{subfig:ablation_b}
    \end{subfigure}\hfill
    \begin{subfigure}[t]{0.29\textwidth}
        \centering
          \includegraphics[width=\linewidth]{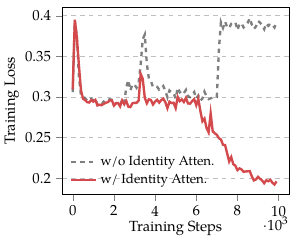}
        \caption{\footnotesize Training loss of 
        \ours{} w and w/o Identity Attention.}
        \label{subfig:albation_c}
    \end{subfigure}\hfill
    \caption{{\bf \ours{} Training and Inference Performance.} (a)  \ours{} achieves optimal generation performance with few inference steps using the DDIM solver; (b) As training progresses, \ours{} shows improvement in the tasks of both video interpolation and image-to-video. These results are based on a mask ratio ranging from 0.15 to 0.6 for 9-frame generation; and (c) The design of Identity Attention is crucial for stable training convergence in \ours{} during the initial training stage; without it, the model fails to converge.}
    \label{fig:ab_curve}
\end{figure}

\paragraph{Efficiency Analysis.} 
Table~\ref{tab:efficiency} illustrates latency and memory usage across different input resolutions and frame lengths, measured on the same computational platform. When MAR is set to operate symmetrically with the DM with the same inputs, the model cannot fit in the available GPU memory as we increase the resolution and/or number of frames. In contrast, our asymmetric design enables the generation of 12-frame clips at 512 resolution in just a few seconds. The rapid generation process is partially attributed to the DM requiring relatively few inference steps to converge, thanks to the well-structured planning signal it receives, as shown in Figure~\ref{subfig:ablation_a}. Notably, inference speed could be further optimized, as the only acceleration technique we incorporated during our experiments is mixed precision, without employing caching strategies \citep{liu2024faster,zhao2024real}, FSDP, or static compilation of the underlying computational graph. Similarly, memory usage could be further reduced through CPU offloading, sliced attention, sequential VAE inference, \textit{etc.}

\begin{figure}[ht!]
    \centering
    \includegraphics[width=0.495\linewidth]{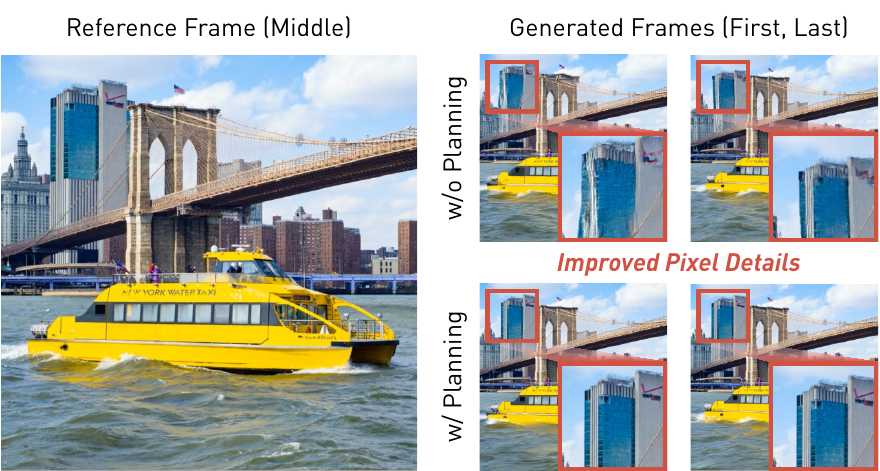}\hfill
    \includegraphics[width=0.495\linewidth]{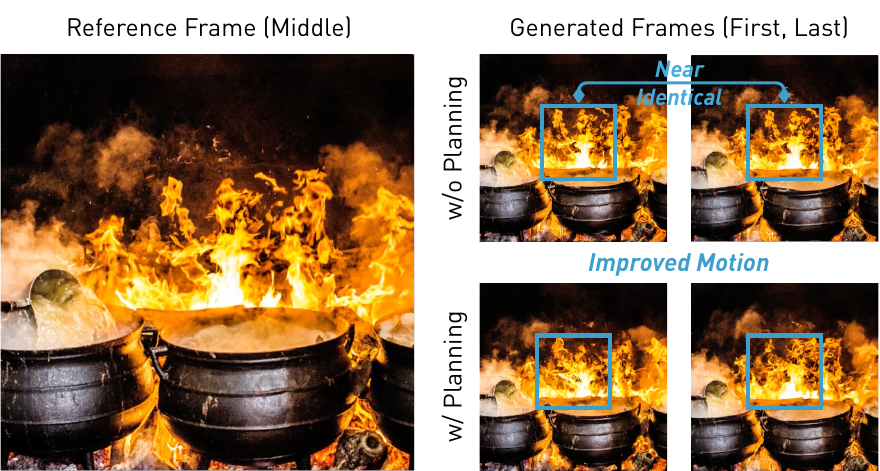}
    \caption{{\bf \ours{}'s generations with and without the planning model.} Here we show video frames generated when conditioning on the middle frame. Without MAR's planning signal, DM generates degraded motion, such as pixel distortions (highlighted in \textcolor{myred}{{\bf red}}, left) or incorrect motions (highlighted in \textcolor{myblue}{{\bf blue}}, right).}
    \label{fig:ab_MAR}
\end{figure}

\paragraph{Explaining MAR's Planning Signal.} We provide an intuitive explanation of MAR's role in \ours{}. During training, a learnable token is used to randomly replace MAR to support CFG \citep{ho2022classifier}, allowing DM to generate videos independently. We visualize the results of \ours{} with and without planning signals. As shown in Figure~\ref{fig:ab_MAR}, without the planning model, DM can still produce meaningful frames but, as expected, lacks ``global planning.'' In Figure~\ref{fig:ab_MAR} (Left), DM moves objects in different directions, causing distortion in the building, which suggests a weaker or non-existing prior model of how objects move. Similarly, in Figure~\ref{fig:ab_MAR} (Right), DM fails to accurately predict the movement of the fire. In contrast, incorporating the planning signal addresses these visual flaws. These results indicate that MAR's planning signal effectively hints how elements should move, ensuring long-term coherence in the generated video.

\paragraph{From Video Interpolation to Image-To-Video Generation.}
Our training recipe follows the philosophy of transitioning from video interpolation to image animation. Herein, we empirically demonstrate that these two tasks are related, validating the soundness of our pipeline. As shown in Figure~\ref{subfig:ablation_b}, we track the performance of \ours{} on both video interpolation and image animation during a training phase aimed at scaling the resolution from 256 to 512. This stage marks the first point during training where the model successfully performs both tasks simultaneously. We observe a promising consistency between the performance of image animation and video interpolation, providing solid evidence that these tasks do not hinder each other. Furthermore, with a carefully tuned mask ratio, the model can be trained in a unified manner to efficiently achieve both tasks.

\paragraph{Impact of Identity Attention.} We explore the effectiveness of Identity Attention in handling our specific data format, which integrates both reference frames and noised frames into a single sequence. As illustrated in Figure~\ref{subfig:albation_c}, we track the training trajectory in the early stages of the DM generation model. We recognize that this type of input can lead to unstable training, particularly when starting from scratch, as the differences between reference frames are difficult to discern. However, the proposed Identity Attention mechanism mitigates this instability. The decrease in training loss observed after 6K steps is attributed to the use of a warm-up learning rate, where the learning rate is intentionally kept low during the initial steps.

\subsection{Results on Video Interpolation}
\label{sec:vi}

\begin{table}[ht!]
\centering
\caption{{\bf Performance of zero-shot video interpolation on VIDIM-Bench.} The reported results are taken directly from VIDIM \citep{jain2024video}. AMT, RIFE, and FILM are single-inference methods, while LDMVFI, VIDIM, and our approach are based on diffusion models with multiple inference steps. MidF-SSIM and MidF-LPIPS represent the SSIM and LPIPS scores, respectively, for the middle frame. For \ours{}-512, we downscale the generated videos to 256 resolution for a fair comparison.}
\renewcommand{\arraystretch}{1.0}
\setlength\tabcolsep{6.2pt}
\footnotesize
\begin{tabular}{lcccccccc}
\toprule
\multirow{2}{*}{Method} & \multicolumn{4}{c}{DAVIS-7}                                          & \multicolumn{4}{c}{UCF101-7}                                          \\\cmidrule(r){2-5} \cmidrule(l){6-9}
                        & MidF-SSIM       & MidF-LPIPS      & FID            & \cellcolor{metabg}FVD             & MidF-SSIM       & MidF-LPIPS      & FID            & \cellcolor{metabg}FVD             \\ \midrule
AMT \citep{li2023amt}                    & 0.4853          & 0.2865          & 34.65          & \cellcolor{metabg}234.50           & 0.7903          & 0.1691          & 31.60          & \cellcolor{metabg}344.50          \\
RIFE \citep{huang2022real}                   & 0.4546          & 0.2954          & 23.98          & \cellcolor{metabg}240.04          & 0.7769          & 0.1564          & 18.72          & \cellcolor{metabg}323.80          \\
FILM \citep{reda2022film}                   & 0.4718          & 0.3048          & 30.16          & \cellcolor{metabg}214.80           & 0.7869 & 0.1620          & 26.06          & \cellcolor{metabg}328.20          \\ \midrule
LDMVFI \citep{danier2024ldmvfi}                  & 0.4175          & 0.2765 & 22.10          & \cellcolor{metabg}245.02          & 0.7712          & \textbf{0.1564} & \textbf{18.09} & \cellcolor{metabg}316.30          \\
VIDIM  \citep{jain2024video}           & 0.4221          & 0.2986          & 28.06          & \cellcolor{metabg} 199.32          & 0.6880          & 0.1768          & 34.48          & \cellcolor{metabg}278.00          \\ \midrule
\ours{}-S/ST-256               & 0.4249          & 0.3654          & 49.21          & \cellcolor{metabg}224.07          & 0.7654          & 0.2480          & 45.85          & \cellcolor{metabg}258.08          \\
\ours{}-L/ST-256              & 0.4959 & 0.2768          & \textbf{20.64} & \cellcolor{metabg}102.87 & 0.7734          & 0.2213          & 28.85        & \cellcolor{metabg}\textbf{197.69} \\ \midrule
\ours{}-S/ST-512               & 0.5017        & 0.3193          & 25.92         & \cellcolor{metabg}138.86          & \textbf{0.7960}          & 0.2315          & 30.24          & \cellcolor{metabg}205.71          \\
\ours{}-L/ST-512              & \textbf{0.5314} & \textbf{0.2736}          & 20.76 & \cellcolor{metabg}\textbf{99.05} & 0.7814          & 0.2347          & 30.08        & \cellcolor{metabg}204.20 \\ \midrule 
\ours{}-L/T-512              & 0.5085 & 0.3083          & 25.30 & \cellcolor{metabg} 117.13 & 0.7893          & 0.2270          & 30.72        & \cellcolor{metabg} 198.94 \\
\bottomrule
\end{tabular}
\label{table:VI}
\end{table}

We compare \ours{} with the existing methods on the VIDIM benchmark \citep{jain2024video} for video interpolation, where the task is to generate 7 frames between a starting and an ending conditional frames. As shown in Table~\ref{table:VI}, \ours{} achieves competitive performance among different evaluation metrics. In particular, it is widely acknowledged that generative models often underperform in reconstruction metrics, with blurrier images often scoring higher despite receiving lower ratings from human observers \citep{sahak2023denoising, watson2022novel, jain2024video,saharia2022image}. 
We also study a sample that is exemplifying of this statement in the Appendix \ref{app:sec:recon_metrics}. Therefore, we place greater emphasis on the generative metric, FVD, where \ours{} outperforms competitors and achieves state-of-the-art performance. Notably, \ours{}-L/T employs an asymmetric attention mechanism, where the planning model utilizes spatio-temporal attention, while the generation model relies on temporal attention. Compared to the model that uses spatio-temporal attention for both models (\ours{}-L/ST), the results suggest that the asymmetric attention mechanism does not significantly affect performance, achieving a satisfactory trade-off between efficiency and quality. We provide additional visualizations in Appendix \ref{app:sec:vis_vi} and the supplementary materials.

\subsection{Results on Image-to-Video Generation}

In this section, we evaluate our model's single-image-to-video generation capabilities in comparison with other methods using the VBench dataset  \citep{huang2023vbench}. As shown in Table~\ref{table:vbench}, our method performs competitively, especially in terms of latency, despite incorporating expensive spatio-temporal attention. For fairness, latency is calculated with the same resolution. In this study, we focus on validating the soundness of our proposed roadmap, only considering the initial pre-training stage rather than delving into post-training techniques. As a result, we do not incorporate additional conditional signals such as language instructions or motion score guidance. Therefore, direct comparisons on video quality, particularly in relation to dynamic degree, are not entirely fair. However, we fully report these numbers for reference.

\begin{table}[ht!]
\caption{{\bf Image-to-Video Performance on VBench.} The reported results of baseline methods are sourced from VBench \citep{huang2023vbench}. For fair latency comparison, we standardize the input size to [512$\times$512] for low and medium resolutions, and [768$\times$768] for high resolution cases across all methods. All other metrics were collected using the original resolutions reported in the first column.}
\label{table:vbench}
\scriptsize
\setlength\tabcolsep{4.5pt}
\begin{tabular}{lcccccccc}
\toprule
Method           & \begin{tabular}[c]{@{}c@{}}Frame \\ Resolution    \end{tabular}          & \begin{tabular}[c]{@{}c@{}}Image-based \\ Pre-training\end{tabular} & \begin{tabular}[c]{@{}c@{}}Latency \\ (s/frame)\end{tabular} & \begin{tabular}[c]{@{}c@{}}I2V Sub.\\ Con\end{tabular} & \begin{tabular}[c]{@{}c@{}}I2V Back\\ Con.\end{tabular} & {\color[HTML]{C0C0C0} \begin{tabular}[c]{@{}c@{}}Video Quality\\ (w/ D.D.)\end{tabular}} & \begin{tabular}[c]{@{}c@{}}Video Quality\\ (w/o D.D.)\end{tabular} & \begin{tabular}[c]{@{}c@{}}Vbench \\ Avg.\end{tabular} \\
\midrule 
\rowcolor[HTML]{EFEFEF} 
 \multicolumn{9}{c}{Low and Medium Resolution} \\ \midrule
ConsistI2V \citep{ren2024consisti2v}       &  [256$\times$256]                      & \Checkmark   &  7.63    & 95.82                                                  & 95.95                                                   & {\color[HTML]{C0C0C0} 78.87}                                                             & 85.74                                                              & 88.27       \\
DynamicCrafter \citep{xing2023dynamicrafter}    & [256$\times$256]                     & \Checkmark                                                                   &    -     & 97.05                                                  & 97.56                                                   & {\color[HTML]{C0C0C0} 80.18}                                                             & 85.00                                                              & 88.07       \\
DynamicCrafter \citep{xing2023dynamicrafter}    &  [512$\times$320]                      & \Checkmark                                                                   & 4.88  & 97.21                                                  & 97.40                                                   & {\color[HTML]{C0C0C0} 81.63}                                                             & 85.39                                                              & 88.37       \\
SEINE  \citep{chen2023seine}            &  [512$\times$320]                  & \Checkmark                                                                   & -  & 96.57                                                  & 96.80                                                   & {\color[HTML]{C0C0C0} 79.49}                                                             & 85.71                                                              & 88.45       \\
VideoCrafter \citep{chen2024videocrafter2}     & [512$\times$320]              & \Checkmark                                                                   &  9.43   & 91.17                                                  & 91.31                                                   & {\color[HTML]{C0C0C0} 81.34}                                                             & 87.55                                                              & 88.47       \\
SEINE \citep{chen2023seine}             & [512$\times$512]                     & \Checkmark                                                                   &     5.13    & 97.15                                                  & 96.94                                                   & {\color[HTML]{C0C0C0} 80.58}                                                             & 87.13                                                              & 89.61       \\
Animate-Anything \citep{dai2023fine}  & [512$\times$512]                & \Checkmark                                                                   &  1.58  & 98.76                                                  & \textbf{98.58}                                                   & \cellcolor[HTML]{FFFFFF}{\color[HTML]{C0C0C0} 81.21}                                     & \cellcolor[HTML]{FFFFFF}\textbf{88.84}                                      & \textbf{91.30}       \\ \midrule
\ours{}-L/ST-9  & [512$\times$512]                     & \XSolidBrush                                                                   &   2.24        & 98.64                                                  & 97.12                                                   & {\color[HTML]{C0C0C0} 80.84 }                                                                  & 88.22                                                              & 90.64       \\
\ours{}-S/ST-9   & [512$\times$512]                     & \XSolidBrush                                                                   &  2.24        & \textbf{99.04}                                                  & 97.23                                                   & {\color[HTML]{C0C0C0} 81.00}                                                                  & 88.59                                                              & 90.98       \\
\rowcolor{metabg} 
\ours{}-L/T-17  & [512$\times$512]                     & \XSolidBrush                                                                   &   0.48      & 98.23                                                  & 97.01                                                   & {\color[HTML]{C0C0C0} 80.25 }                                                                  & 87.68                                                              & 90.16       \\
\rowcolor{metabg} 
\ours{}-S/T-17   & [512$\times$512]                     & \XSolidBrush                                                                   &   \textbf{0.46}      & 98.76                                                  & 97.18                                                   & {\color[HTML]{C0C0C0} 80.56 }                                                                  & 88.17                                                              & 90.62       \\
\midrule
\rowcolor[HTML]{EFEFEF} 
 \multicolumn{9}{c}{High Resolution} \\ \midrule
SVD-XT-1.0 \citep{blattmann2023stable}       & [1024$\times$576]              & \Checkmark                                                                   &  2.19       & 97.52                                                  & 97.63                                                   & {\color[HTML]{C0C0C0} 82.79}                                                             & 86.54                                                              & 89.30       \\
SVD-XT-1.1  \citep{blattmann2023stable}      & [1024$\times$576]                & \Checkmark                                                                   &  2.19      & 97.51                                                  & 97.62                                                   & {\color[HTML]{C0C0C0} 82.23}                                                             & 86.66                                                              & 89.38       \\
I2VGen-XL \citep{2023i2vgenxl}       & [1280$\times$720]                & \Checkmark                                                                   &  6.01  & 96.48                                                  & 96.83                                                   & {\color[HTML]{C0C0C0} 81.17}                                                             & 87.02                                                              & 89.43       \\
DynamiCrafter \citep{xing2023dynamicrafter}     & [1024$\times$576]                     & \Checkmark                                                                   &  7.13  & 98.17                                                  & \textbf{98.60}                                                   & {\color[HTML]{C0C0C0} 82.52}                                                             & 87.31                                                              & 90.08       \\ \midrule
\rowcolor{metabg} 
\ours{}-L/T-17   & [768$\times$768] & \XSolidBrush                                                                   &  1.01       &  98.34                                                 &  96.63                                                  & {\color[HTML]{C0C0C0} 80.88 }                                                                  &      88.22                                                       &   90.54   \\
\rowcolor{metabg} 
\ours{}-S/T-17   & [768$\times$768] & \XSolidBrush                                                                   &   \textbf{0.98}  &     98.77                                              &   96.78                                                & {\color[HTML]{C0C0C0} 81.29}                                                                  &           88.68                                                   &    90.95    \\
\rowcolor{metabg} 
MARDini-L/T-17   & [1024$\times$1024] & \XSolidBrush                                                                   &   -  &     98.61                                            &  96.34                                               & {\color[HTML]{C0C0C0} 81.35}                                                                  &           88.69                                                   &    90.89    \\

\rowcolor{metabg} 
MARDini-S/T-17   & [1024$\times$1024] & \XSolidBrush                                                                   &   -  &     \textbf{98.78}                                              &  96.46                                               & {\color[HTML]{C0C0C0} 81.74}                                                                  &           \textbf{88.97}                                                   &    \textbf{91.13}    \\
\bottomrule
\end{tabular}
\end{table}

We also report the results on the benchmark without the motion score (referred to as Dynamic Degree in VBench). All evaluation metrics are detailed in Appendix \ref{app:sec:bench}. The empirical study shows \ours{}'s strong potential, performing on par with other existing methods across several metrics while exhibiting higher efficiency and requiring no generative image pre-training. Interestingly, we observe that \ours{}-S marginally outperforms \ours{}-L on some evaluation metrics. We speculate that this is due to \ours{}-L requiring more training time to accommodate higher-resolution data. Nonetheless, we observe clear advantages in scaling the MAR model size, as \ours{}-L outperforms in video interpolation and generates image-to-video results that better align with physical principles. A list of generated video samples is provided in the supplementary for further reference.

\subsection{Additional Applications}
\label{sec:discuss}
In this section, we explore some of \ours{}'s additional intriguing capabilities and applications. While we did not conduct rigorous ablation studies or quantitative comparisons, this serves as an initial exploration, highlighting potential directions for future research.

\begin{figure}[ht!]
   \centering
   \small
   \renewcommand{\arraystretch}{1.0}
   \setlength{\tabcolsep}{0.2em}
   \resizebox{.99\textwidth}{!}{
       \begin{tabular}{*{6}{C{0.166\linewidth}}}
       \multicolumn{2}{c}{Reference Frames (First, Last)} & \multicolumn{4}{c}{Generated Frames} \\
      \cmidrule(lr){1-2} \cmidrule(lr){3-6}
       \includegraphics[width=\linewidth]{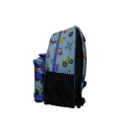} &
       \includegraphics[width=\linewidth]{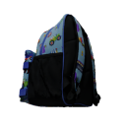} &
       \includegraphics[width=\linewidth]{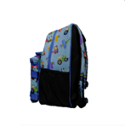} &
       \includegraphics[width=\linewidth]{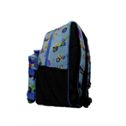} &
       \includegraphics[width=\linewidth]{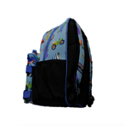} &
       \includegraphics[width=\linewidth]{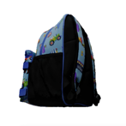} 
        \\
       \includegraphics[width=\linewidth]{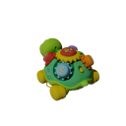} &
       \includegraphics[width=\linewidth]{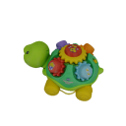} &
       \includegraphics[width=\linewidth]{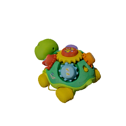} &
       \includegraphics[width=\linewidth]{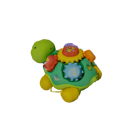} &
       \includegraphics[width=\linewidth]{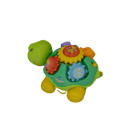} &
       \includegraphics[width=\linewidth]{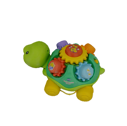} 
        \\
    \end{tabular}
    }
    \caption{{\bf Visualization of novel view synthesis conditioned on the two views.} Starting with two views of an object, \ours{} generates the intermediate ``frames'', effectively creating novel views. Notably, \ours{} is trained without any 3D data but still manages to capture spatial information through video. The data used for this task is sourced from publicly available research datasets \citep{downs2022google}.}
    \label{fig:3d}
\end{figure}

\paragraph{Zero-Shot 3D Novel View Synthesis}
We demonstrate \ours{}'s strong potential for 3D novel view synthesis. Although trained solely on video data, \ours{} exhibits a preliminary level of spatial understanding, suggesting its potential for 3D applications. In Figure~\ref{fig:3d}, two views of a fixed object serve as the first and last reference frames, while intermediate frames are generated, as similar to our video interpolation task. The model effectively generates convincing 3D-consistent views, highlighting its promising potential for 3D generation. Notably, no camera control signals are used, and we will explore \ours{} on 3D data with better control in the future work.

\begin{figure}[ht!]
   \centering
   \small
   \renewcommand{\arraystretch}{1.0}
   \setlength{\tabcolsep}{0.2em}
   \resizebox{.99\textwidth}{!}{
       \begin{tabular}{*{2}{C{0.166\textwidth}}*{4}{C{0.166\textwidth}}}
       \multicolumn{2}{c}{Reference Frames} & \multicolumn{4}{c}{Generated Frames} \\
       \cmidrule(lr){1-2} \cmidrule(lr){3-6}
       \includegraphics[width=\linewidth]{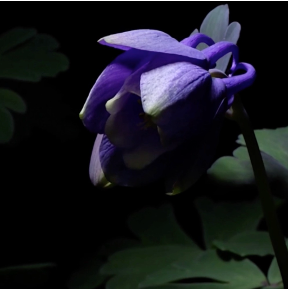} &
       \includegraphics[width=\linewidth]{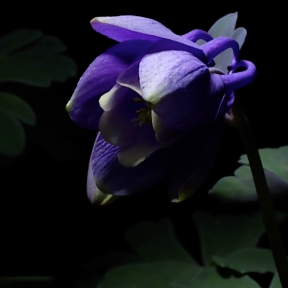} &
       \includegraphics[width=\linewidth]{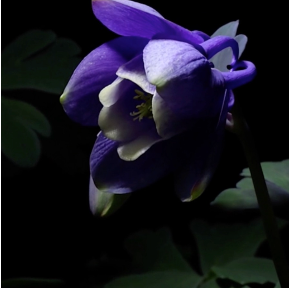} &
       \includegraphics[width=\linewidth]{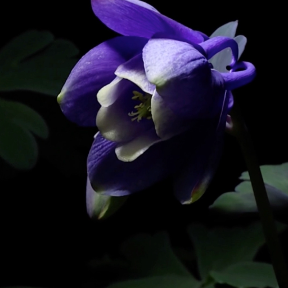} &
       \includegraphics[width=\linewidth]{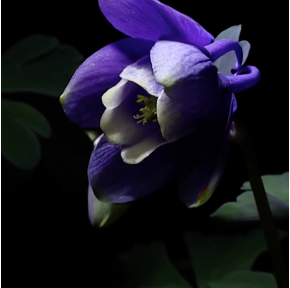} &
       \includegraphics[width=\linewidth]{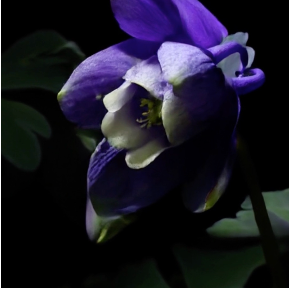} 
        \\
       \includegraphics[width=\linewidth]{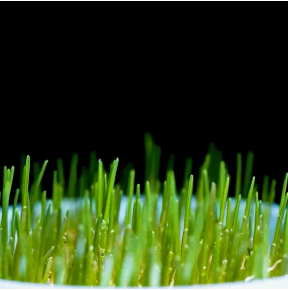} &
       \includegraphics[width=\linewidth]{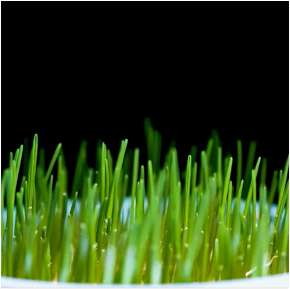} &
       \includegraphics[width=\linewidth]{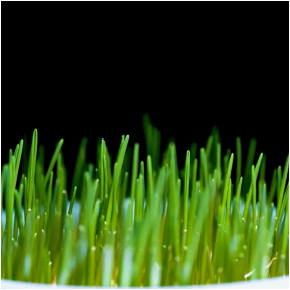} &
       \includegraphics[width=\linewidth]{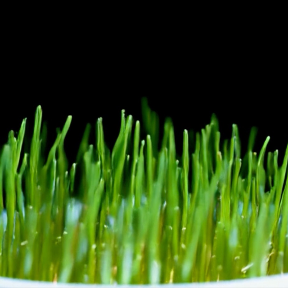} &
       \includegraphics[width=\linewidth]{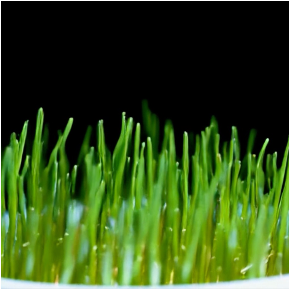} &
       \includegraphics[width=\linewidth]{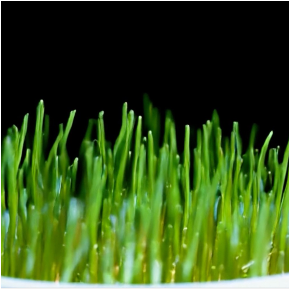} 
        \\
    \end{tabular}
    }
    \caption{{\bf Visualization of Video Expansion.} The model is conditioned on a sequence of 16 consecutive frames to predict the subsequent 12 frames. The video data used for visualization is sourced from publicly available research dataset \citep{nan2024openvid}.}
    \label{fig:scene}
\end{figure}

\paragraph{Video Expansion} \ours{} integrates many of MAR's advantages, including the support for video expansion, where the conditional input is a set of frames rather than a single image. In this setup, motion information is implicitly embedded in the input. As shown in Figure~\ref{fig:scene}, \ours{} can effectively predict video sequences based on the provided motion cues (e.g., flower blooming, grass growing).

\paragraph{(Hierarchical) Auto-Regressive Generation}
By utilizing MAR for high-level planning, \ours{} also supports auto-regressive inference, generating more frames beyond the one defined in the training stage. We demonstrate this through hierarchical auto-regressive generation: starting with a given video, we segment it into multiple clips, expand each clip segment, and treat the expanded clip segment as the new video for recursive video interpolation. In Figure~\ref{fig:auto} (in Appendix), we provide an example where,
starting with 4 images, \ours{} with a 32-frame window size auto-regressively expands them into a 128-frame slow-motion video (32$\times$ expansion). This illustrates that our model is not limited by the training window size, highlighting its potential for long-range video generation.

\section{Related Work}

\paragraph{Auto-Regressive Model in Visual Generation.} 
Auto-regressive (AR) models \citep{gers2000learning,hochreiter1997long,schmidhuber2015deep} have proven effective in natural language modeling \citep{brown2020language,achiam2023gpt,dubey2024llama,team2023gemini}. 
To adapt this scalable modeling strategy for image and video generation, recent approaches \citep{yu2023language,chang2022maskgit,li2023mage,yu2023magvit,chang2023muse,yu2023magvit} replace causal attention in AR with bidirectional attention, allowing for better capture of dense relationships in visual space.

Many studies \citep{yu2023scaling,chang2023muse,team2024chameleon,xie2024showo} validate the scalability of this approach.
To align with the training recipes from LLMs, these studies adopt discrete visual representations, using image tokenizers \citep{esser2021taming,yu2021vector,van2017neural} to quantize continuous pixel values into discrete representations. However, \cite{li2024autoregressive,ramesh2021zero,razavi2019vqvae2} argue that this strategy suffers from unstable training and may limit model capacity due to the inherently continuous nature of visual data. This inspires recent works \citep{li2024autoregressive,zhou2024transfusion} to shift towards continuous latent spaces for masked auto-regressive models to address these limitations.

We follow this trajectory but diverges in two ways: i) We highlight the importance of mask ratios, which were fixed in earlier works \cite{li2024autoregressive}. By dynamically adjusting them with a progressive training strategy, we improve both model scalability and stability. ii) We propose an asymmetric input resolution design, allowing MAR to be effectively trained with full-resolution inputs.

\paragraph{Diffusion Model for Video Generation.} In recent years, diffusion models \citep{ho2020denoising,neal2001annealed,jarzynski1997equilibrium} have become a leading approach for both image and video generation \citep{rombach2022high,dhariwal2021diffusion,ramesh2022hierarchical,chen2024gentron,saharia2022photorealistic,brooksvideo,dai2023emu,girdhar2023emu,menapace2024snap,kondratyuk2023videopoet,cong2023flatten}. 
These models conceptualize the generation process as gradually refining a real sample from Gaussian noise, demonstrating significant scalability and stable training.

In this paper, we offer two key insights into video generation: i) Previous methods \citep{wu2023tune,ho2022imagen,zhang2023show,blattmann2023align,wang2023modelscope,girdhar2023emu,gao2024lumina,cong2023flatten} often first pre-train an image generative model, and then fine-tune it for video generation, or they require joint training for both tasks \citep{chen2024gentron,esser2023structure}. While multi-stage pre-training on diverse inputs can be beneficial, video generation is often limited by the success of image-based pre-training, which typically serves as a secondary task. This paper proposes an alternative: training video generation models from scratch with progressively increasing task complexity. ii) Previous research \citep{girdhar2023emu, wang2023modelscope, chen2024gentron,blattmann2023align} has predominantly employed temporal attention mechanisms to capture temporal dependencies, mainly due to the high computational and memory costs associated with spatio-temporal attention. However, in alignment with previous work \citep{blattmann2023align,gao2024lumina} suggesting that spatio-temporal attention enables superior video modelling, we propose an amortized strategy that makes spatio-temporal attention computationally feasible, even at high resolutions. 

\paragraph{Asymmetric Neural Networks.}
This paper also relates to asymmetric neural architectures, widely used in neural networks since the 1990s \citep{schmidhuber1992learning,schmidhuber1992learningcontrol}. In computer vision, to achieve high-resolution generation, many studies \citep{podell2023sdxl,pernias2023wurstchen,saharia2022photorealistic,li2024autoregressive,jain2024video,kang2023scaling} employ a common strategy: a model generates low-resolution/quality samples, followed by another model that performs super-resolution \citep{kang2023scaling}, refinement \citep{podell2023sdxl}, or interpolation \citep{wang2024magicvideo} to enhance the generation quality. In discriminative video models, asymmetric training strategies have been used for temporal segmentation models, where the full temporal extension does not fit the available GPU memory~\cite{xu2021low}. Since computational costs are distributed across stages, this approach is well-supported by existing computational platforms.
Building on this trajectory but extending beyond it, we propose a novel design that partitions the model into two distinct models: a planning model and a generation model. The planning model, containing the majority of the model's parameters, is trained auto-regressively at a low resolution to generate conditional signals without producing visual outputs. These signals are then processed by the lightweight generation model, which converts them into high-resolution visual outputs using a diffusion process. 

Unlike the traditional auto-regressive diffusion model \citep{li2024autoregressive}, which still faces high computational costs as resolution increases, we use cross-attention as an information pathway to connect asymmetric resolution input for more efficient training/inference.

\section{Limitations and Future Works}
\label{app:sec:limitation}

\paragraph{Post Training.}
The primary goal of this paper is to demonstrate the feasibility and effectiveness of combining masked auto-regressive (MAR) models with diffusion models (DM) for video generation. Consequently, we allocated the majority of our computational resources to the pre-training stage, placing less emphasis on post-training, despite its recognized importance in generative models \citep{dai2023emu,dubey2024llama,touvron2023llama}. Post-training will be a top priority in our future work, focusing on enhancing long-term planning, improving motion quality, and achieving higher resolutions.

\paragraph{Improved Conditional Signals.} A significant contribution of this work is the exploration of training a video generation model without relying on generative image pre-training. However, this approach presents a trade-off: \ours{} is not inherently equipped with a text encoder for processing language-based instructions. To conserve computational resources and quickly validate the feasibility of our method, we intentionally excluded commonly used conditional signals, such as text embeddings and motion scores. Encouraged by the initial success of our model, we plan to incorporate these conditional signals into \ours{} in our future updates to broaden its range of applications.

\section{Conclusion}
We have introduced a new family of generative models for video, i.e.,~\ours{}, based on auto-regressive diffusion, wherein a large planning model offers powerful conditioning to a much smaller diffusion model. Our design philosophy considers efficiency from model conception, and so our heaviest model component is only executed once at lower resolution inputs, whereas our generative module focuses on fine-grained details at the frame level, reconciling high-level conditioning and image details. Our model is unique in that it leverages a masked auto-regressive loss directly at the frame level. \ours{}~is afforded with multiple generative capabilities from a single model, e.g., long-term video interpolation, video expansion, and image animation. Our investigation shows that our modeling strategy is powerful enough to obtain competitive results on various interpolation and animation benchmarks, while doing it at a lower computational needs than counterparts with comparable parameter size.

\section*{Acknowledgements}
The authors thank Mingchen Zhuge, Jinheng Xie, Yuren Cong, Kam Woh Ng, Aditya Patel, and Jinjie Mai for their valuable suggestions and contributions to the paper review. Haozhe Liu and Jürgen Schmidhuber were supported by funding from the King Abdullah University of Science and Technology (KAUST) - Center of Excellence for Generative AI under award number 5940 and the SDAIA-KAUST Center of Excellence in Data Science and Artificial Intelligence.

\section*{Ethics Statement}
This paper explores the theoretical foundations of neural architecture design for video generation, rather than being tied to specific commercial applications. Consequently, the potential negative impacts of \ours{} align with those of other video generation models and do not pose unique risks that require special consideration. Importantly, unlike previous models trained on web-scale data, which may raise concerns about data copyright, \ours{} is exclusively trained on a licensed Shutterstock dataset, without having such conflicts. 
\clearpage
\newpage
\bibliographystyle{assets/plainnat}
\bibliography{paper}

\clearpage
\newpage
\beginappendix
\section{Reconstruction metrics in Video Interpolation.}
\label{app:sec:recon_metrics}

\begin{figure}[ht!]
     \begin{subfigure}[t]{0.32\textwidth}
        \centering
          \includegraphics[width=\linewidth]{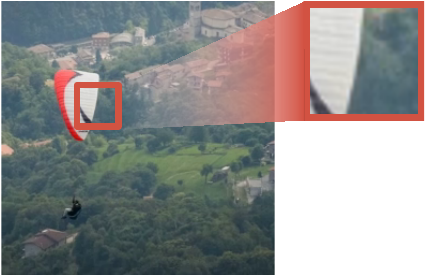}
        \caption{\footnotesize Original Video}
    \end{subfigure}\hfill
    \begin{subfigure}[t]{0.32\textwidth}
        \centering
        \includegraphics[width=\linewidth]{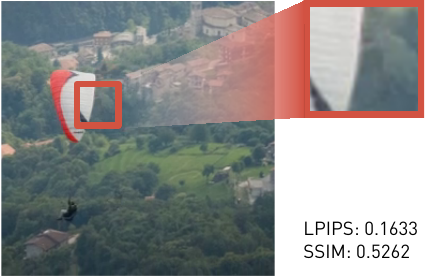}
        \caption{\footnotesize LDMVFI}
    \end{subfigure}\hfill
    \begin{subfigure}[t]{0.32\textwidth}
        \centering
          \includegraphics[width=\linewidth]{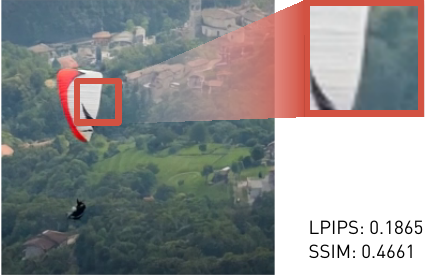}
        \caption{\footnotesize Ours}
    \end{subfigure}
    \caption{\footnotesize{\bf Failure case of reconstruction metrics (SSIM, LPIPS) in video interpolation.} We visualize two generated frames together with their corresponding ground-truth frames. While the frames generated by \ours{} are sharper than competitors, their corresponding reconstruction scores are worse.}
    \label{fig:recon}
\end{figure}

In Figure~\ref{fig:recon},  it appears that blurrier images sometimes receive higher reconstruction error scores.

\section{\ours{} Training Strategies}
\label{app:sec:training_detials}
\ours{} is trained on the Shutterstock video dataset with 34 million videos, using 256 H100 GPUs with a distributed MAST scheduler \citep{choudhury2024mast}. We use the AdamW optimizer for each stage with a $1.4 \times 10^{-4}$ learning rate and cosine learning rate scheduler. We adapt our batch size based on the resolution and the frame count to maximize GPU utility. For example, at $[256\times256]$ resolution with 9 frames, the batch size is 1024, processing 9K frames per iteration; at $[512\times512]$ resolution with 9 frames, the batch size is 720, processing 6480 frames per iteration. During inference, we set the classifier-free guidance (CFG)\citep{ho2022classifier} scale as 2.5 for the image-to-video task with the noise solver DDIM \citep{song2020denoising}, and we directly remove classifier-free guidance for video interpolation as it is redundant. FSDP \citep{zhao2023pytorch} and activation checkpointing \citep{zhao2023pytorch} are enabled to further save GPU memory. We do not include dynamic resolution training in our main training stages, as it slows down training. Instead, we find that after convergence, fine-tuning the model for a few steps (10K-20K) with dynamic resolutions enables it to quickly support this capabilities.

\section{Visualization of Video Interpolation}
\label{app:sec:vis_vi}
\begin{figure}[ht!]
   \centering
   \footnotesize
   \renewcommand{\arraystretch}{1.0}
   \setlength{\tabcolsep}{0.2em}
   \resizebox{.99\textwidth}{!}{
       \begin{tabular}{*{7}{C{0.135\linewidth}}}
        & & \multicolumn{4}{c}{Generated Frames (Middle)} & \\
       \cmidrule(lr){3-6}
       \multicolumn{2}{c}{Reference Frames (First, Last)} & FILM & LDMVFI & VIDIM & Ours & Ground-Truth \\
       \includegraphics[width=\linewidth]{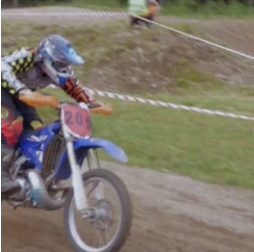} &
       \includegraphics[width=\linewidth]{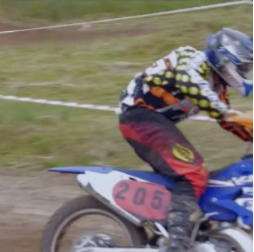} &
       \includegraphics[width=\linewidth]{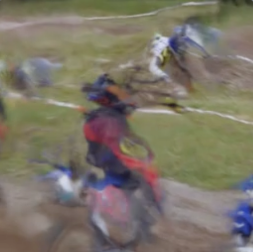} &
       \includegraphics[width=\linewidth]{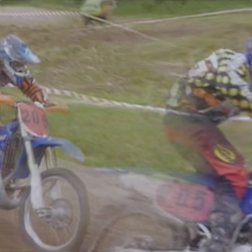} &
       \includegraphics[width=\linewidth]{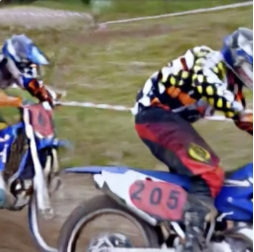} &
       \includegraphics[width=\linewidth]{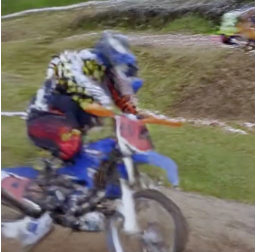} &
       \includegraphics[width=\linewidth]{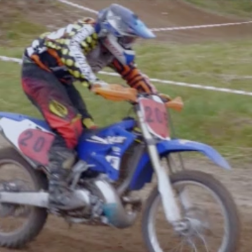} 
        \\
    \end{tabular}
    }
     \caption{\footnotesize{\bf Visualization of video interpolation methods conditioned on the first and last frames.} We present the generated frames from FILM \citep{reda2022film}, LDMVFI \citep{danier2024ldmvfi}, VIDIM \citep{jain2024video}, and \ours{}. The comparison results for these methods are sourced from \cite{jain2024video}. We have included additional samples in the supplementary materials. }
    \label{fig:vi_inter}
\end{figure}

In Figure~\ref{fig:vi_inter}, we provide visualization results that demonstrate the superiority of \ours{} in large motion modelling, compared to FILM \citep{reda2022film}, LDMVFI \citep{danier2024ldmvfi}, and VIDIM \citep{jain2024video}.

\newpage
\section{Benchmarks} 
\label{app:sec:bench}
We evaluate the interpolation performance on VIDIM-Bench \citep{jain2024video} and assess image animation performance on VBench \citep{huang2023vbench}. 

For VIDIM-Bench, the task involves generating seven intermediate frames, with the first and last frames provided as conditions. The dataset includes approximately $400$ videos from both DAVIS \citep{pont20172017} and UCF-101 \citep{Soomro2012UCF101AD}. We use FVD \citep{unterthiner2018towards} and FID \citep{heusel2017gans} as generation metrics, while adopting SSIM \citep{wang2004image} and LPIPS \citep{zhang2018unreasonable} as reconstruction metrics. Notably, we evaluate the middle (5th) frame for reconstruction metrics, as it presents the greatest challenge due to its distance from the reference frames. 

For VBench, we utilize the official dataset to assess the model across several metrics: I2V-Subject Consistency, I2V-Background Consistency, and video quality. The video quality evaluation considers dimensions such as Subject Consistency, Background Consistency, Smoothness, Aesthetic Score, Imaging Quality, Temporal Flickering, and Dynamic Degree. Given that our model lacks text supervision, we omit the evaluation for video-text camera motion. Furthermore, since our model is pre-trained without incorporating dynamic degree guidance (known as motion score/strength), it is not directly comparable with other models in this respect. Therefore, we additionally report video quality by averaging all the dimensions except for Dynamic Degree and provide the VBench average score derived from I2V-Subject Consistency, I2V-Background Consistency, and the video quality dimensions (excluding dynamic degree). For the latency analysis, we ensure fairness by using the same computational platform: a single Nvidia A100 80G GPU. All implementations are based on their official code without any engineering optimizations. For \ours{}, we simply employ \texttt{bf16} mixed precision to enhance computational efficiency. To account for variations in frame number and resolution, all results are normalized by frame count and evaluated at a consistent resolution of either [512 $\times$ 512] or [768 $\times$ 768].

\begin{figure}[!ht]
    \centering
    \includegraphics[width=0.85\linewidth]{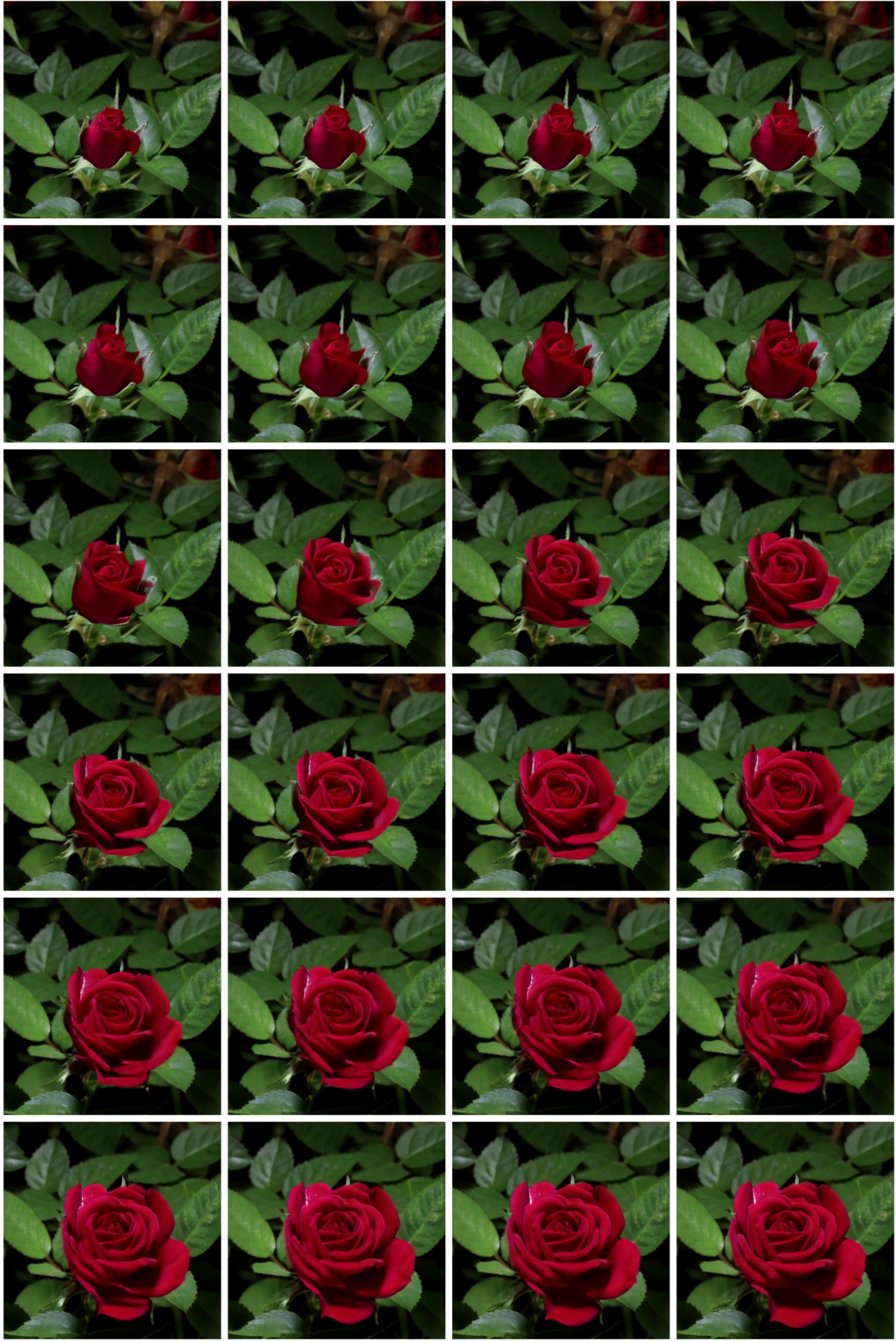} 
    \caption{{\bf Visualization of \ours{} using hierarchical auto-regressive generation.} Starting with an initial 4 frames, \ours{} auto-regressively generates a complete 128-frame video, demonstrating its capability to extend beyond the training window size (32 frames here).}
    \label{fig:auto}
\end{figure}

\end{document}